\documentclass[runningheads]{llncs}
\usepackage{booktabs}

\usepackage[T1]{fontenc}
\usepackage{graphicx}
\usepackage{amsmath}
\usepackage{amssymb}
\usepackage{color}
\usepackage{multirow}
\usepackage{array}
\usepackage{placeins}
\usepackage{tablefootnote}
\usepackage{hyperref}
\usepackage{subcaption}
\usepackage{cite}

\begin{document}
\title{Counterfactual Understanding via Retrieval-aware Multimodal Modeling for Time-to-Event Survival Prediction}
\titlerunning{Retrieval-aware Multimodal Time-to-Event Survival Prediction}
\author{Ha-Anh Hoang Nguyen\thanks{Equal contribution}\inst{1}\orcidID{0009-0008-3643-3170} \and
Tri-Duc Phan Le\protect\footnotemark[1]\inst{1}\orcidID{0009-0003-2998-6408} \and
Duc-Hoang Pham\inst{1}\orcidID{0009-0007-6847-384X} \and
Huy-Son Nguyen\inst{2}\orcidID{0009-0006-4616-0976} \and
Cam-Van Thi Nguyen\inst{1}\orcidID{0009-0001-9675-2105} \and
Duc-Trong Le\inst{1}\orcidID{0000-0003-4621-8956} \and
Hoang-Quynh Le\thanks{Corresponding author.}\inst{1}\orcidID{0000-0002-1778-0600}}
\authorrunning{H.A.H. Nguyen et al.}
\institute{University of Engineering and Technology,\\Hanoi Vietnam National University, Vietnam \\
\email{\{23020513, 23020048, 22021200, vanntc, trongld, lhquynh\}@vnu.edu.vn}
\and
Delft University of Technology, the Netherlands \\
\email{\{h.s.nguyen@tudelft.nl\}}}

\maketitle              % typeset the header of the contribution
\vspace{-10pt}
\begin{abstract}
This paper tackles the problem of time-to-event counterfactual survival prediction, aiming to optimize individualized survival outcomes in the presence of heterogeneity and censored data. We propose CURE, a framework that advances counterfactual survival modeling via comprehensive multimodal embedding and latent subgroup retrieval. CURE integrates clinical, paraclinical, demographic, and multi-omics information, which are aligned and fused through cross-attention mechanisms. Complex multi-omics signals can be adaptively refined using a mixture-of-experts architecture, emphasizing the most informative omics components. Building upon this representation, CURE implicitly retrieves patient-specific latent subgroups that capture both baseline survival dynamics and treatment-dependent variations. Experimental results on METABRIC and TCGA-LUAD datasets demonstrate that proposed CURE model consistently outperforms strong baselines in survival analysis, evaluated using the Time-dependent Concordance Index ($C^{td}$) and Integrated Brier Score (IBS). These findings highlight the potential of CURE to enhance multimodal understanding and serve as a foundation for future treatment recommendation models.
All code and related resources are publicly available to facilitate the reproducibility\footnote{\url{https://github.com/L2R-UET/CURE}}.

\end{abstract}
\vspace{-10pt}
\section{Introduction}
\textit{Time-to-event survival prediction} is a fundamental task in medical prognosis, aiming to estimate the time until a specific clinical event (such as disease recurrence, progression, or death) occurs. Chen et al.~\cite{chen2024introduction} showed that survival analysis has long studied time-to-event outcomes, and recent methods emphasize personalized prediction of such times.
Despite extensive progress in statistical and machine learning–based survival models, several challenges remain unresolved, including the handling of censored data, patient heterogeneity, and the need for interpretable, individualized predictions \cite{chen2024introduction,lee2018deephit}.

Clinical decision-making inherently involves evaluating alternative or no-treatment scenarios that cannot be simultaneously observed~\cite{jamil2025gascade}. In practice, for any given patient, only one treatment outcome is realized, while outcomes under other possible interventions remain unobserved. \textit{Counterfactual} reasoning provides a principled framework to address this limitation by estimating these unobserved outcomes and reasoning about what would have happened under different therapeutic choices \cite{haugh2023counterfactual}. This enables the estimation of personalized treatment effects, i.e., how a patient’s survival outcome would differ across alternative or no-treatment scenarios. Counterfactual modeling establishes a robust foundation for treatment-specific personalized survival prediction and interpretable decision support.
Building upon this idea, \textit{personalized survival modeling} aims to account for patient heterogeneity, as treatment efficacy often varies significantly across subpopulations. Most existing survival prediction approaches rely on the Cox proportional hazards model and its extensions \cite{abadi2014cox,katzman2018deepsurv,lee2018deephit,kvamme2019time}. They typically assume homogeneous treatment effects and may overlook latent variability in patient responses. Recent advances, namely the Cox Mixture with Heterogeneous Effects (CMHE) \cite{nagpal2022counterfactual}, provide a flexible framework for discovering counterfactual phenotypes—latent patient subgroups exhibiting distinct survival behaviors and treatment responses under alternative interventions. 

%However, accurately modeling these individualized effects remains highly challenging. 
%Review: The meaning multimodality deviates a bit from what people in IR consider as multimodal "text, vision, audio etc
In healthcare, multimodality denotes the integration of heterogeneous patient data sources. Deep learning models can jointly process multiple biomedical and contextual data types, including multi-omics, imaging, clinical records, text, and social data~\cite{acosta2022multimodal}. Moreover, a patient’s response to treatment is rarely determined by shallow signals such as demographic attributes alone. While clinical and paraclinical indicators provide valuable contextual information, they are often insufficient to capture the underlying biological mechanisms driving treatment response. Richer modalities, i.e. \textit{multi-omics} data describing genomic, transcriptomic, or proteomic profiles can reveal deep, personalized biological factors influencing survival outcomes~\cite{wiegrebe2024deep}. Yet integrating such heterogeneous information sources introduces additional complexity due to their high dimensionality, modality gaps, and varying levels of noise.

To address these challenges, we propose \textbf{CURE} (\textbf{C}ounterfactual \textbf{U}nderstan\-ding via \textbf{R}etrieval-aware Multimodal Modeling for Time-to-\textbf{E}vent Survival Prediction). 
\textit{Unlike traditional retrieval-based approaches} that rely on explicit patient matching, C\textit{URE performs latent, retrieval-aware multimodal mechanism}, emulating the retrieval process in a differentiable latent space rather than depending on direct similarity search.
Explicit similarity search is often unreliable in multimodal clinical data because features are sparse, censored, noisy, and partially missing.
Instead, CURE performs latent retrieval, implicitly aligning patients within a learned counterfactual space that encodes treatment–response mechanisms rather than raw feature similarity.
This formulation enables more faithful modeling of causal treatment–outcome relationships and provides three key advantages: (i) generalization to unseen patients by reasoning over learned subpopulation priors instead of fixed neighbor indices; (ii) modeling of continuous clinical heterogeneity via soft, probabilistic subgroup assignment; and (iii) improved causal interpretability through latent treatment–response archetypes.

Our main contributions are as follows:
\begin{itemize}
\vspace{-3pt}
\item We introduce a retrieval-aware survival prediction model that performs implicit retrieval over counterfactual treatment-response subgroups, enabling individualized and interpretable survival estimation. By reasoning over learned latent subpopulations instead of explicit patient-level similarity, the model achieves more robust generalization across heterogeneous and noisy multimodal cohorts.
\item We design a cross-attentive multimodal encoder that effectively aligns clinical, paraclinical, and multi-omics data into a unified patient representation, in which a hierarchical mixture-of-experts module emphasizes the most informative omics components for each individual, filtering out irrelevant signals for personalized survival prediction.
\item We conduct extensive experiments on two publicly available datasets and demonstrate significant improvements over all competing methods across multiple evaluation metrics.
\end{itemize}

\vspace{-10pt}
\section{Related Work}
% Traditional Survival Models
%Classical survival analysis primarily relies on statistical models such as the Cox proportional hazards (CoxPH) model~\cite{cox1972regression} (1972), which assumes a log-linear relationship between covariates and the hazard function under the proportional hazards assumption. Some extensions includes: a neural network replacing Cox's linear risk function that can model the complex nonlinear correlation of patients' features and risks~\cite{faraggi1995neural,katzman2018deepsurv} and an application of Random Forests to censored data that estimates the cumulative hazard function~\cite{ishwaran2008random}.
%However, these models are still using the proportional hazard assumption and therefore cannot model the risks changing over time, which is a more common case in the real world.
A variety of survival models have shown promising advances in learning complex relationships between covariates and time-to-event outcomes. 
\textit{Classical survival analysis} primarily relies on statistical models such as the Cox proportional hazards model and its extensions~\cite{cox1972regression,faraggi1995neural,katzman2018deepsurv,ishwaran2008random}. They use the proportional hazards assumption and therefore cannot model risks that change over time in real-world data.
% More advanced survival models
\textit{Recent deep survival models} leverage the assumption by predicting different risks across multiple points in time~\cite{zhong2021deep}. DeepHit~\cite{lee2018deephit}, Cox-Time~\cite{kvamme2019time}, extend the Cox framework with deep neural networks that capture how individual risk evolves over time based on complex, nonlinear covariates and hazard interactions. SA-DGNet~\cite{yang2024deep} even employs self-attention gating-based network for more accurate prediction.  Despite their success, most of these methods still unable to effectively tackle the treatment effect aspect, which is essential for any healthcare problem, as they primarily focus on predicting survival trend. 

%Counterfactual and Treatment
Recent approaches have introduced \textit{counterfactual reasoning and causal representation learning} for survival data. Counterfactual Mixture-of-Hazard Experts (CMHE)~\cite{nagpal2022counterfactual} extends this line by modeling a patient’s survival with and without treatment through a mixture-gates formulation. While CMHE represents a major step toward survival prediction, it relies mainly on clinical and treatment variables and does not leverage the rich biological context from multi-omics and other high-dimensional modalities.

%Retrieval-enhanced causal inference / survival modeling (suggested by reviewer)
Other studies have further refined \textit{causal survival modeling} by focusing on metrics like the restricted mean lifetime to quantify treatment differences \cite{chen2001causal}. Additionally, hybrid approaches, such as combining propensity score weighting with classification trees, have been developed to enhance causal effect estimation in time-to-event data \cite{linden2018estimating}. Explicit patient retrieval from electronic medical records has been used for clinical exploration, but such approaches are not designed for counterfactual survival analysis~\cite{6061403}. While these methods offer robust statistical frameworks, they typically do not provide the end-to-end integration of high-dimensional, multimodal omics signals utilized by our CURE framework.

%CURE
Existing deep survival models primarily learn population-level risks, while counterfactual approaches such as CMHE move toward personalization but often overlook multimodal and omics-level signals.
Our proposed CURE framework bridges these gaps by integrating heterogeneous multimodal information through cross-attention–based fusion and mixture-of-experts adaptation, while leveraging implicit latent subgroup retrieval to jointly capture baseline survival patterns and treatment-dependent heterogeneity, thereby enabling personalized and counterfactual survival prediction.
% %-------------------------
% A variety of survival models have shown promising advances in learning complex relationships between covariates and time-to-event outcomes, including DeepSurv \cite{katzman2018deepsurv} (2018), DeepHit \cite{lee2018deephit} (2018) and SA-DGNet \cite{yang2024deep} (2024). However, all  of these works solely rely on unimodal inputs, which does not capture the complex multimodal interactions. Furthermore, omics data are not being integrated, missing the opportunity for more accurate survival prediction and treatment effect estimation.

\vspace{-10pt}
\section{Methodology}
\vspace{-5pt}
\begin{figure}[t]
\centering
\includegraphics[width=\textwidth]{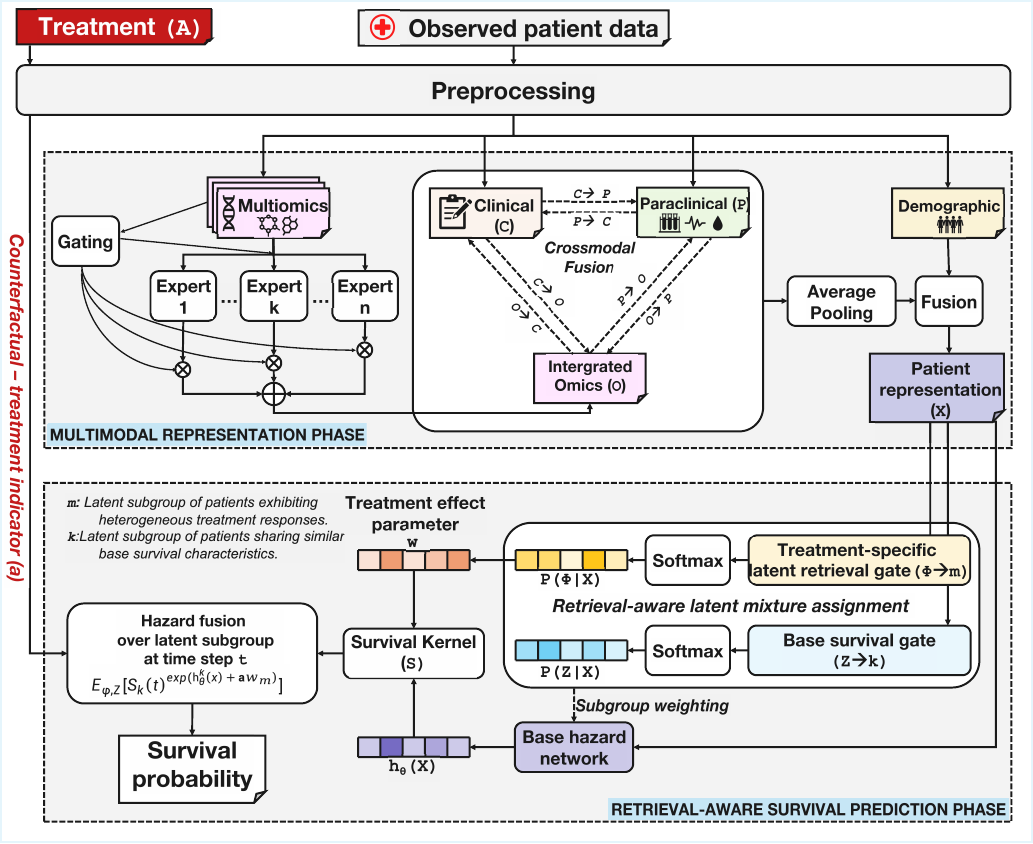}
\caption{Overall architecture of the proposed CURE model.}
\label{fig:model}
\vspace{-10pt}
\end{figure}
To enable effective and personalized estimation in multimodal, high-dimensional, and partially censored settings, we propose \textbf{CURE}, a two-phase framework for counterfactual survival prediction.  
As illustrated in Figure~\ref{fig:model}, CURE consists of two main phases:  
(1) the \textit{multimodal representation phase}, which learns comprehensive patient embeddings through cross-modal fusion, and  
(2) the \textit{retrieval-aware survival prediction phase}, 
which employs latent mixture retrieval to infer subgroup-specific hazard functions, 
capturing treatment heterogeneity and enabling personalized survival estimation.
%Each phase is described in detail in the following subsections.

\textbf{Preliminaries.}
The \textit{time-to-event survival problem} with right-censored data aims to estimate event times under incomplete observations.  
For each patient $i$, let $T_i$ denote the true event time and $C_i$ the censoring time.  
The observed time is defined as $Y_i = \min(T_i, C_i)$, with an event indicator $\delta_i = \mathbb{I}[T_i \le C_i]$.  
Given multimodal patient covariates $X_i = \{x_i^{(r)}\}_{r=1}^R$, where $r$ denotes the modality index (e.g., clinical, paraclinical, demographic, or omics features) and each omics modality further contains multiple submodalities $X_i^{(O)} = \{x_i^s\}_{s=1}^S$, where $s$ indexes individual omics types, along with the treatment assignment $A_i$, the goal is to estimate the conditional survival function:
\begin{equation}
    S(t \mid X_i, A_i) = P(T_i > t \mid X_i, A_i),
\end{equation}
which represents the probability that patient $i$ survives beyond time $t$ under treatment $A_i$.  
%Learning $S(t \mid X, A)$ requires handling multimodal and censored data,  
while ensuring both discrimination %(ranking accuracy via the Concordance Index, $C^{td}$)  
and \textit{calibration}. $X_i$.% (risk estimation accuracy via the Integrated Brier Score, IBS).

\textbf{Assumptions.}
Following prior counterfactual survival modeling work~\cite{nagpal2022counterfactual}, we assume \textit{conditional unconfoundedness}, i.e., that the treatment assignment $A_i$ is independent of the potential outcomes given the observed patient's covariates $X_i$. The distribution of time to event $T$ is assumed to be independent conditional on the covariates and treatment assignment. Regarding the latent subgroups, they are also assumed to be independent given covariates and the time-to-event distributions in each subgroup obey proportional hazards. Under these assumptions, we are able to make subgroup-level counterfactual inference.
\vspace{-10pt}
\subsection{Multimodal Representation Phase}
CURE takes as input diverse patient information, including multi-omics, clinical, paraclinical, and demographic data, each appropriately preprocessed for its respective dataset. To integrate these information, the multimodal representation phase employs a multimodal architecture with cross-attention–based fusion.
%All modalities are aligned at the patient (sample) level using unique sample identifiers, and survival outcomes are defined by the time-to-event (T) and event indicator (E) variables.

\textbf{a. Omics Representation via Mixture-of-Experts.}
Multi-omics data can include genomics, transcriptomics, proteomics, metabolomics, etc. Such data are often highly complex and high-dimensional, with variability across datasets; even within the same dataset, not all patients have complete omics information, and among available data, not all features are equally informative in a given context. Therefore, a mixture-of-experts approach~\cite{wu2024multi} is employed to select the most relevant omics information and produce a single, unified representation vector.

Preprocessing and normalization involve \texttt{log} or \texttt{z-score} normalization, filtering of low-variance or rare features, and encoding. Subsequently, each omics matrix is compressed through a linear bottleneck layer $(d_{\text{pre}} \ll d_{\text{omics}})$ as:
\begin{equation}
    h^{(s)} = \phi ( W_{\text{pre}}^{(s)} X^{(s)} + b_{\text{pre}}^{(s)} )
\end{equation}
where $h^{(s)}$ denotes the low dimensional latent embedding of the $s$-th omics modality after linear compression, $\phi(\cdot)$ denotes a composite transformation consisting of layer normalization, ReLU activation, and dropout regularization, $W_{\text{pre}}^{(s)} \in \mathbb{R}^{d_{\text{pre}}\times d_{\text{omics}}}$ is the projection matrix that maps the original omics feature space to the lower-dimensional bottleneck subpace and $b^{(s)}_{\text{pre}} \in \mathbb{R}^{d_{\text{pre}}}$ is the bias vector added for each modality to correct for systematic shifts in feature distributions.

This process reduces redundancy and stabilizes feature variance while preserving survival-relevant molecular patterns. 
The resulting low-rank embeddings are further refined by omic-specific Mixture of Experts (MoE) encoders, where a gating network adaptively selects the most relevant experts for each sample \cite{wu2024multi,nguyen2025multi}.  
This joint bottleneck–MoE design enables scalable, noise-robust modeling of high-dimensional omics data, producing compact and discriminative embeddings.
To effectively capture molecular heterogeneity, each omics modality is processed through an MoE module that partitions the feature space into multiple specialized subregions, as follows:
\begin{equation}
    z^{(s)} = \sum_{e=1}^{E} \alpha_{e}^{(s)} f_{e}^{(s)} (h^{(s)})
\end{equation}
where $z^{(s)}$ denotes the refined latent representation of the $s$-th omics modality after expert aggregation and $f_e^{s}(\cdot)$ represents the $e$-th expert network responsible for modeling a particular subspace of molecular variation.
Besides, $\alpha^{(s)}_e = \mathrm{softmax}(G^{(s)}(h^{(s)}))_e$ are sample-dependent gating weights that determine the contribution of each expert. $G^{(s)}(\cdot)$ is the gating network that maps the input representation $h^{(s)}$ to a vector of expert relevance scores in $\mathbb{R}^E$. The gating mechanism dynamically selects and weights experts based on the latent structure of each sample, allowing adaptive specialization across heterogeneous omics signals.

Through this adaptive specialization, the MoE enhances the representation of nonlinear relationships and latent substructures within omics data, yielding biologically meaningful embeddings that coherently with clinical and paraclinical features during subsequent multimodal cross-attention steps.

\textbf{b. Cross-Modal Fusion via Pairwise Attention.} To capture pairwise relationships between heterogeneous modalities, we employ a pairwise cross-attention fusion mechanism inspired by the Transformer architecture. 

For modality $r$ with input vector $X_r \in \mathbb{R}^{N \times d_r}$, where $N$ is the batch size and $d_r$ is the feature dimension of modality $r$, we produce an embedding $\mathbf{Z}_r \in \mathbb{R}^{N \times D}$:
\begin{equation}
    \mathbf{Z}_r = \textbf{MLP}_r(X_r),
\end{equation}
where each ${MLP}_m(\cdot)$ consists of two linear layers with ReLU activation; $D$ is the shared embedding dimension across all modalities.
 
Let $\mathbf{C}, \mathbf{P}, \mathbf{O} \in \mathbb{R}^{N \times D}$ denote the embeddings of \textit{Clinical}, \textit{Paraclinical}, and \textit{Omics} modalities, respectively.
For each modality, we compute its cross-attention to the other two modalities.  
For example, when the Clinical modality attends to the Paraclinical one, the query, key, and value projections are given by:
\begin{align}
\mathbf{Q}_{CP} &= \mathbf{C}\mathbf{W}_Q^{CP}, \quad
\mathbf{K}_{CP} = \mathbf{P}\mathbf{W}_K^{CP}, \quad
\mathbf{V}_{CP} = \mathbf{P}\mathbf{W}_V^{CP},
\end{align}
where $\mathbf{W}_Q^{CP}, \mathbf{W}_K^{CP}, \mathbf{W}_V^{CP} \in \mathbb{R}^{D \times D}$ are trainable parameters.  

The cross-attention output is then computed as:
\begin{align}
\mathbf{A}_{CP} &= \mathrm{softmax}\!\left( \frac{\mathbf{Q}_{CP}\mathbf{K}_{CP}^\top}{\sqrt{d_k}} \right), \\
\mathbf{Z}_{CP} &= \mathbf{A}_{CP}\mathbf{V}_{CP},
\end{align}
where $d_k = D /h$, $h$ is the number of attention heads, $\mathbf{A}_{CP}$ is the attention weight matrix and $\mathbf{Z}_{CP}$ is the cross-attended output. 

We employ analogous pairwise attention computations which are performed for all modality pairs $\{\mathbf{Z}_{CO}, \mathbf{Z}_{PC}, \mathbf{Z}_{PO}, \mathbf{Z}_{OC}, \mathbf{Z}_{OP}\}$. The six pairwise-attended representations are averaged to produce a unified fused embedding $\mathbf{Z}_{\text{fused}}$:
\begin{align}
\mathbf{Z}_{\text{fused}} = \frac{1}{6} \sum_{(r,n) \in \mathcal{P}} \mathbf{Z}_{rn}, 
\end{align}
where $\quad 
\mathcal{P} = \{(C,P), (C,O), (P,C), (P,O), (O,C), (O,P)\}$. 

As commonly adopted in prior studies~\cite{tran2025multifaceted}, the fused embedding is then concatenated with the normalized demographic (\texttt{D}) features, which serve as static global descriptors, to form the comprehensive patient representation $\mathbf{X}$, as follows:
\begin{equation}
    \mathbf{X} = [\mathbf{Z}_{\text{fused}} \mathbin\Vert \texttt{D}]
\end{equation}

\vspace{-10pt}
\subsection{Retrieval-aware Survival Prediction Phase}
%Nhắc lại 1 lần nữa tại sao chỉ aware mà ko làm retrieve thật ko reviewer lại hỏi.
This phase is retrieval-aware rather than retrieval-dependent, performing implicit latent retrieval through probabilistic subgroup assignment instead of explicit patient search.
Such a design captures counterfactual treatment–response structures and generalizes better to unseen or censored multimodal data.

\textbf{a. Retrieval-aware Mixture Assignment.}
First, a patient representation $\mathbf{x} \in \mathbf{X}$ is passed through two feed-forward networks $g_k$ and $g_m$, followed by softmax operations to produce the treatment-specific retrieval distribution $P(Z \mid \mathbf{x})$ and the baseline-survival subgroup distribution $P(\Phi \mid \mathbf{x})$:
\begin{equation}
\mathbf{v}_k = g_k(\mathbf{x}) \in \mathbb{R}^K,\quad P(Z=k \mid \mathbf{x}) = \text{Softmax}_k(\mathbf{v}_k)
\end{equation}
\begin{equation}
\mathbf{u}_m = g_m(\mathbf{x}) \in \mathbb{R}^M,\quad P(\Phi=m \mid \mathbf{x}) = \text{Softmax}_m(\mathbf{u}_m)
\end{equation}
where $K$ and $M$ denote the numbers of treatment-response and baseline-survival subgroups, respectively.

Under the assumption of conditional independence given $\mathbf{x}$, these two distributions jointly parameterize a retrieval-aware latent mixture over expert pairs $(k, m)$, which serves as subgroup weights determining the relative contribution of each survival expert to the final hazard and survival predictions:
\begin{equation}
    \mathrm{P}(Z=k, \Phi=m \mid \mathbf{x}) = \mathrm{P}(Z=k \mid \mathbf{x}) \mathrm{P}(\Phi=m \mid \mathbf{x})
\end{equation}

\textbf{b. Base Hazard Network.}
We pass the latent baseline survival subgroup gate $\mathbf{v}_k$ through a baseline survival network $f_{\theta}$ and obtain the group-specific hazard functions, depicted as follow:
\begin{equation}
    h_\theta(\mathbf{x}) = f_{\theta}(\mathbf{v}_k) = [h_\theta^{(1)}(\mathbf{x}), h_\theta^{(2)}(\mathbf{x}), \dots, h_\theta^{(K)}(\mathbf{x})]
\end{equation}
where each component $h_\theta^{(k)}(\mathbf{x})$ captures the baseline survival risk pattern of the patient $\mathbf{x}$ in subgroup $k$.

\textbf{c. Survival Kernel.}
Inspired by the Cox Proportional Hazards \cite{cox1972regression}, which is commonly used in modeling survival with censored outcomes, the conditional hazard rate of a patient is derived as: 
\begin{equation}
    \lambda(t \mid \mathbf{x}) = \lambda_0(t) \exp(h_\theta(\mathbf{x}))
\end{equation}
where $\lambda_0(t)$ denotes the baseline hazard rate shared across the population.

Taken from that idea, we assign each pair of subgroup $(k, m)$ a baseline hazard rate $\lambda^m_k(t)$. From there we define the base survival $S^{m}_{k}(t)$ (survival probability at time $t$ for subgroup pair $(k, m)$). In continuous terms, this kernel is derived from the hazard rate as follows:
 \begin{equation}
     S_k^{m}(t) = \exp\left( - \int_0^t \lambda_k^{m}(u) \, du \right)
 \end{equation}

 Then the survival kernel of patient $\textbf{x}$ with treatment $\textbf{a}$ in subgroup pair $(k, m)$ is defined as: 
 \begin{equation}
     S^{m}_{k}(t \mid \textbf{x}, \textbf{a})= S^m_k(t) ^{exp(h^k_{\theta}(\mathbf{x}) + \mathbf{a} \omega_m)}
 \end{equation}
 where $\mathbf{a}$ is the treatment indicator (1 for treated, 0 for control), and $\omega_m$ denotes the treatment-effect parameter associated with retrieval subgroup $m$. 
 
These subgroup-specific kernels are later aggregated through the posterior mixture weights derived in the previous phase, yielding the final retrieval-aware hazard and survival functions.

\textbf{d. Hazard Fusion over Latent Subgroups at Time Step \texttt{t}.}
Each expert $E^m_k$ in the subgroup pair $(k, m)$ models the time-dependent survival kernel of patient $\mathbf{x}$ and represents the predicted survival probability at time $t$ for patients belonging to latent survival subgroup $k$ and treatment-response subgroup $m$:
\begin{equation}
S^m_k(t \mid \mathbf{x}, \mathbf{a}) = g_{E^m_k}(t, \mathbf{x}, \mathbf{a}),
\end{equation}
where $g_{E^m_k}$ denotes the survival expert network parameterized by $(k, m)$.

The final prediction then aggregates over all experts through a retrieval-aware mixture that fuses subgroup-specific hazards and retrieval-based treatment effects into a unified survival representation, enabling the model to capture individualized and counterfactual survival dynamics:
\begin{equation}
\hat{S}(t \mid \mathbf{x}, \mathbf{a})=\sum^K_{k=1}\sum^M_{m=1}\mathrm{P}(Z=k, \Phi = m|\mathbf{x)}S^m_k(t) ^{exp(h^k_{\theta}(\mathbf{x}) + \mathbf{a} \omega_m)}
\label{eq:fuse}
\end{equation}

The latent retrieval gate $\Phi$ encodes the matching probability between each patient and the underlying treatment response subgroups, effectively retrieving the subgroup whose latent characteristics best align with the patient's multimodal profile.

Formally, the subgroup $m$ with the highest retrieval weight $\mathbf{u}_m$ is selected for each patient. The average effect of treatment $\Omega_m$ within each subgroup $m$ is calculated as follows:
\begin{equation}
\Omega_m = 
\frac{1}{  |\Phi_m|}
\sum_{x \in \Phi_m}
\left[
\int_{0}^{t} \hat{S}(t \mid \mathbf{x}, 1)\,dt
-
\int_{0}^{t} \hat{S}(t \mid \mathbf{x}, 0)\,dt
\right]
\end{equation}
where $\hat{S}(t \mid \mathbf{x}, 1)$ and $\hat{S}(t \mid \mathbf{x}, 0)$ represent the predicted survival functions under treatment and control, respectively.

As a result, we can infer useful information on which patient subtypes derive the most or least benefit from treatment.

\vspace{-10pt}
\section{Experiments and Results}
\vspace{-5pt}
% experimental setup, results and analyses of research questions..
\subsection{Experimental Setup}

% \paragraph

{\textbf{Dataset.}} We evaluate CURE on the Molecular Taxonomy of Breast Cancer International Consortium (METABRIC)~\cite{curtis2012genomic}, and The Cancer Genome Atlas Lung Adenocarcinoma (TCGA - LUAD)~\cite{cancer2014comprehensive} datasets\footnote{METABRIC and TCGA-LUAD are available at \url{https://www.cbioportal.org/}.}. 
Their overall characteristics are presented in Table~\ref{tab:data}. While METABRIC includes nearly four times more patients, TCGA-LUAD provides richer omics information but exhibits a shorter mean survival time and higher censoring ratio, indicating more heterogeneous and challenging data.

\begin{table}[h]
\caption{Data statistics.}
\resizebox{\textwidth}{!}{
\begin{tabular}{l|c|ccc|cc|*{4}{>{\centering\arraybackslash}m{0.8cm}}}
\hline
 & \textbf{Size} 
 & \multicolumn{3}{c|}{\textbf{Duration$^*$}} 
 & \multicolumn{2}{c|}{\textbf{Event}} 
 & \multicolumn{4}{c}{\textbf{\begin{tabular}[c]{@{}c@{}}Modality \\ feature dimension$^+$\end{tabular}}} \\ \cline{3-11}
\textbf{} 
 &  
 & \textbf{Min} & \textbf{Mean} & \textbf{Max} 
 & \textbf{Uncensored} & \textbf{Censored} 
 & \textbf{O} & \textbf{C} & \textbf{P} & \textbf{D} \\ \hline
\textit{METABRIC} 
 & 1,980 
 & 0 & 125 & 355 
 & 1,143 & 837 
 & 56K+ & 15 & 41 & 3 \\ 
\textit{TCGA–LUAD} 
 & 505 
 & 0 & 29.7 & 238 
 & 182 & 323 
 & 60K+ & 11 & 10 & 14 \\ \hline
\multicolumn{11}{r}{
\begin{tabular}[c]{@{}r@{}}
$^*$\textit{Measured in months. $^+$O: Omics, C: Clinical, P: Paraclinical, D: Demographics.}
\end{tabular}}
\end{tabular}
}
\label{tab:data}
\vspace{-10pt}
\end{table}

\textbf{Hyperparameter Settings.}
For each omics modality, the linear bottleneck dimension is $256$.
The MoE adapter employs $4$ experts with top-$2$ routing and an auxiliary coefficient $\lambda_{\text{aux}}{=}0.01$ to prevent expert collapse.
Training uses \texttt{AdamW} (learning rate $5\times10^{-4}$, weight decay $1\times10^{-5}$, batch size $128$) for up to $200$ epochs.
Each modality embedding is processed by a $2$-layer Transformer encoder with $4$ attention heads and dropout $0.1$.
For survival prediction and subgroup retrieval, the number of latent subgroups is set to two or three for treatment-effect groups, and in range of $[1;5]$ for base survival groups.
Models are optimized with \texttt{Adam}~\cite{kingma2014adam} (learning rate $0.01$) for $20$ epochs and batch size $100$.

\textbf{Computational Efficiency.} We use T4 GPU on Google Colab to train the model and it takes approximately 5 hours on training and over 1 minute for inference.

\textbf{Metrics.} We adopt two widely used metrics in survival analysis:
$C^{td}$ (\textit{Time-dependent Concordance Index})~\cite{antolini2005time} evaluates the model’s ability to correctly rank comparable patient pairs by survival risk; higher values indicate better discrimination.
\texttt{IBS} (\textit{Integrated Brier Score}) measures the mean squared error between predicted survival probabilities and observed outcomes over time; lower values indicate better calibration and accuracy.
%Longer version
% \textbf{$C^{td}$} (Time-dependent Concordance Index ($C^{td}$)~\cite{antolini2005time}) measures the model’s ability to correctly rank survival times between comparable patient pairs.
% It estimates the probability that, for any two comparable patients, the one with a shorter observed survival time is assigned a higher predicted risk score. A higher $C^{td}$ therefore indicates better discriminative ability of the model.
% %estimates the probability that observations $i$ and $j$ are concordant (true order is preserved), given they are comparable:
% %\[C^{td} = P\{\hat{S}(T_i \, | \, \mathbf{x}_i) < \hat{S}(T_j \, | \, \mathbf{x}_j)\, | \,T_i < T_j, E_i = 1\}\]
% \textbf{IBS} (Integrated Brier Score) extends the Brier Score by integrating prediction errors over the entire follow-up period.
% It quantifies the mean squared difference between the predicted survival probabilities and the actual survival outcomes across time.
% A lower IBS value indicates better overall calibration and predictive accuracy.

%is extended from the Brier Score \cite{graf1999assessment} - the MSE around the probabilistic prediction at a certain time horizon:
% \[\text{BS}(t) = \mathbf{E}[||\mathbf{1}\{T > t\} - \hat{P} (T > t \, | \, X)||^2_2]\]
% The IBS is then defined as follow:
% \[\text{IBS} = \frac{1}{t_2 - t_1} \int_{t_1}^{t_2}\text{BS}(s)ds\]

\textbf{Comparative Models.}
We compared the CURE performance with the following models:
% \vspace{-5pt}
\begin{itemize}
    \item CPH (1972) \cite{cox1972regression}-traditional Cox Proportional Hazards; DeepSurv (2018) \cite{katzman2018deepsurv} - an extended Cox by replacing its linear risk function with a neural network.
    \item RSF (2008) \cite{ishwaran2008random} applied random forests to censored survival data.
    \item DeepHit (2018) \cite{lee2018deephit} and Cox-Time (2019) \cite{kvamme2019time} relaxed the proportional hazards assumption to model risks changing over time.
    \item CMHE (2022) \cite{nagpal2022counterfactual} employed an implicit MoE architecture to learn counterfactual treatment effect.
    \item SA-DGNet (2024) \cite{yang2024deep} introduced a gating-based architecture that adaptively selects informative subnetworks.
\end{itemize}

\begin{table}[t]
\centering
\caption{Overall model performance on METABRIC and TCGA-LUAD datasets.}
\label{tab:survival_results_improve}
\small
\setlength{\tabcolsep}{4pt}
\begin{tabular}{l|cccc|cccc}
\hline
\multirow{2}{*}{\textbf{Model}} & \multicolumn{4}{c|}{\textbf{METABRIC}} & \multicolumn{4}{c}{\textbf{TCGA-LUAD}} \\ %\cline{2-9}
 & \textbf{$C^{td}$}$^{\dagger}$ & $\Delta C^{td\dagger}$ & \textbf{IBS}$^\ddagger$ & $\Delta$IBS$^\ddagger$ &
   \textbf{$C^{td}$}$^{\dagger}$ & $\Delta C^{td\dagger}$ & \textbf{IBS}$^\ddagger$ & $\Delta$IBS$^\ddagger$ \\ \hline
CPH$^*$        & 0.628 & \textit{0.041} & 0.183 & \textit{0.018} & -- & -- & -- & -- \\
Cox-Time$^*$   & 0.662 & \textit{0.007 }& 0.172 & \textit{0.007} & -- & -- & -- & -- \\
RSF$^*$        & 0.652 & \textit{0.017} & 0.176 & \textit{0.011} & -- & -- & -- & -- \\
DeepSurv       & 0.654 & \textit{0.015} & \underline{0.167} & \textit{0.002} & 0.511 & \textit{0.241} & 0.189 & \textit{0.025} \\
DeepHit        & 0.656 & \textit{0.013} & 0.176 & \textit{0.011 }& 0.539 & \textit{0.213} & 0.264 & \textit{0.100} \\
SA-DGNet       & \underline{0.668} & \textit{0.001} & 0.194 & \textit{0.029} & \underline{0.590} & \textit{0.162} & \underline{0.166} & \textit{0.002} \\
CMHE$^+$       & 0.563 & \textit{0.106} & 0.185 & \textit{0.019} & 0.616 & \textit{0.136} & 0.198 & \textit{0.034} \\ \hline
\textbf{CURE}  & \textbf{0.669} & -- & \textbf{0.165} & -- & \textbf{0.752} & -- & \textbf{0.164} & -- \\ \hline
\multicolumn{9}{r}{\begin{tabular}[c]{@{}r@{}}\footnotesize{\textit{Best results are in \textbf{bold}. Second best results are \underline{underlined}.}}\\ $^*$\footnotesize{\textit{Results are reported in Kvamme et al.~\cite{kvamme2019time}. $^+$Re-implemented results.}}\end{tabular}}                                                                            \\
\multicolumn{9}{r}{\begin{tabular}[c]{@{}r@{}}$^\dagger$\footnotesize{\textit{A higher $C^{td}$ denotes better concordance and discriminative performance.}} 
\\ $^\ddagger$\footnotesize{\textit{A lower IBS reflects better calibration and prediction accuracy. }}\\
%\footnotesize{\textit{$\Delta C^{td}$=CURE - Comparative result. $\Delta IBS$ = Comparative result - CURE.}}
\end{tabular}} 
\label{tab:results}
\end{tabular}
\vspace{-20pt}
\end{table}

\vspace{-10pt}
\subsection{Experimental Results}
\vspace{-5pt}
\textbf{a. Overall Performance. }
Table~\ref{tab:results} summarizes the comparative performance of recent survival models.
CURE archives the highest $C^{td}$ and lowest \texttt{IBS} on both datasets, underscoring its effectiveness.% in jointly learning rich multimodal representations and latent treatment–response subgroups for counterfactual survival prediction.
On METABRIC, performance gaps between CURE and competing models are diminutive ($\Delta C^{td}$: $0.01$–$0.106$, $\Delta$IBS: $0.002$–$0.029$).
TCGA-LUAD has smaller size, higher censoring rate, and richer omics modalities. Nevertheless, on TCGA-LUAD, CURE achieves the larger improvement margin ($\Delta C^{td}$: $0.136$–$0.241$, $\Delta$IBS: $0.002$–$0.100$), demonstrating its strong ability to model complex, highly censored multimodal survival data.

% \vspace{-20pt}
% \vspace{-30pt}
% CURE achieves the highest \texttt{$C^{td}$} and lowest \texttt{IBS} on both datasets. 
% On METABRIC, CURE attains a \texttt{$C^{td}$} of 0.669 and an \texttt{IBS} of 0.165, slightly improving over the strongest baseline (SA-DGNet) by 0.001 in $C^{td}$ and -0.029 in IBS. On TCGA-LUAD dataset, CURE yields a substantial improvement—achieving a $C^{td}$ of 0.752 and IBS of 0.164, outperforming SA-DGNet and others by at least 0.162 and -0.002, respectively. 
% All of these altogether highlight the ability of CURE to make advantage of the relationships between the modalities and omics data to produce well-calibrated survival estimation.

\textbf{b. Contribution Analysis of Model Components and Modalities.}
\begin{figure}[h]
\vspace{-10pt}
\centering
\includegraphics[width=\linewidth]{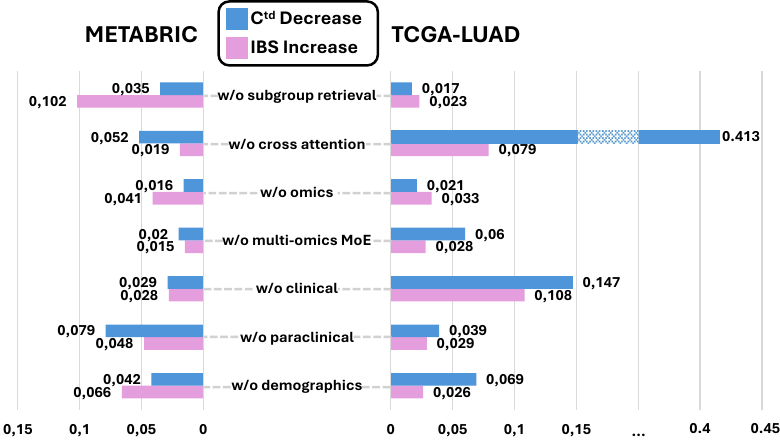}
\caption{Contribution analysis of model components and modalities.}
\label{fig:ablation}
\vspace{-10pt}
\end{figure}

We perform ablation studies by removing each component or modality in turn and evaluating the resulting models on the METABRICS and TCGA-LUAD datasets (Fig.~\ref{fig:ablation}).
%The performance differences from the full model are shown in Fig.~\ref{fig:ablation}.
The results indicate that all components and modalities contribute positively to the overall performance, in terms of the increments in $C^{td}$ and decrements in $IBS$, though the degree of improvement varies across components, modalities, and datasets.
Performance drops on LUAD are relatively larger than on METABRIC, suggesting that component contributions become more evident when the data exhibit higher heterogeneity, censoring, and multimodal complexity.
Removing the cross-attention mechanism leads to a substantial performance drop on both datasets, even larger than removing individual modalities.
Notably, on TCGA-LUAD, eliminating this component results in severe degradation in both $C^{td}$ and \texttt{IBS}% (e.g., $0.413$ drop in $C^{td}$ without cross attention)
, highlighting its key role in modeling cross-modality dependencies.
In contrast, subgroup retrieval shows a stronger contribution on METABRIC%, particularly in terms of \texttt{IBS}, which decreases by $0.102$ when the component is removed
. This effect may stem from the larger sample size of METABRIC, allowing the model to form more distinct latent subgroups in training process.

\textbf{c. Latent Subgroups Analysis.}
\begin{figure}[h]
\vspace{-10pt}
\centering
\includegraphics[width=\linewidth]{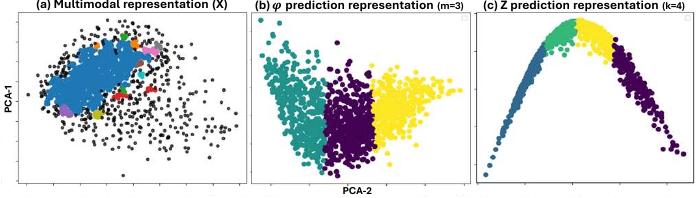}
\caption{Latent subgroups visualization on METABRIC dataset.}
\label{fig:subgroup}
\vspace{-10pt}
\end{figure}

Fig.~\ref{fig:subgroup} visualizes patient representations at three modeling stages on METABRIC dataset using PCA.
In the original multimodal representation space (\texttt{X}), the samples are distributed almost randomly, with a Hopkins statistic $H\approx0.2$, indicating a weak cluster tendency and high heterogeneity among patients. 

The treatment-specific latent retrieval prediction ($\Phi$-prediction) organizes patients into discernible clusters corresponding to latent treatment–response subgroups. By calculating the risk difference (treatment effect) between $A=0$ and $A=1$, we identify three response profiles: \textbf{low responders} (cyan) with negligible effect ($0.0055 \pm 0.0069$), \textbf{moderate responders} (purple) with moderate benefit ($0.0101 \pm 0.0087$), and \textbf{high responders} (yellow) with the largest risk reduction ($0.0277 \pm 0.0200$).

Meanwhile, the survival gating space (\texttt{Z}-prediction) partitions patients into distinct baseline risk strata. Our quantitative analysis of these groups reveals the following risk profiles: \textbf{low risk profiles} (blue) with mean risk of $0.0311 \pm 0.0512$, \textbf{moderate-to-low risk profiles} (green) with mean risk of $0.0725 \pm 0.1130$, \textbf{high risk profiles} (yellow and purple) with mean risk of $0.3597 \pm 0.3766$ and $0.4713 \pm 0.1753$, respectively.

These visualizations confirm that the retrieval-aware latent mixture assignment step enables CURE to disentangle treatment-specific and baseline survival factors in a purely latent manner, without explicit patient matching.

\vspace{-10pt}
\section{Conclusion}
\vspace{-5pt}
This paper presents CURE, a multimodal retrieval-aware framework for predicting counterfactual survival.
CURE includes a retrieval-aware survival prediction module that implicitly retrieves latent treatment-response subgroups, allowing individualized and interpretable estimation of counterfactual survival outcomes.
By reasoning over learned latent subpopulations rather than explicit patient-level similarity, CURE achieves robust generalization across heterogeneous and noisy multimodal cohorts.
In parallel, a multimodal encoder with cross-attention and mixture-of-experts mechanisms effectively aligns clinical, paraclinical, and multi-omics data into a unified representation.
Extensive experiments across METABRIC and TCGA-LUAD datasets demonstrate that CURE achieves superior results compared to established baselines, achieving the highest $C^{td}$ and \texttt{IBS}.
% Ablation analyses further highlight the contribution of model components and modalities.
Thereby, our approach can pave the way for dynamic treatment recommendation by incorporating longitudinal data and graph-based patient representations~\cite{gao2020mgnn} to capture temporal evolution and inter-patient similarity. 
%Review: The cross-attention fusion is described mathematically, but not enough detail is given on how modality gaps, missing data, and feature imbalance are handled in practice during training and inference.
Future work will enhance robustness in real-world clinical settings by addressing modality gaps, missing data, and feature imbalance through modality-aware training strategies such as gradient modulation~\cite{peng2022balanced} or prototypical modal regularization~\cite{fan2023pmr}.

%\textit{Submissions of papers must be at least 6 pages (the sixth page should have at least some content but not necessary to fill it) and at most 12 pages in length plus additional pages for references. The "Theory of Change" section and Appendices count toward the page limit.}

\vspace{-10pt}
\section{Theories of Change}
\vspace{-5pt}
%Section bắt buộc của ECIR IR for good track.
\textbf{Societal Need.}
Selecting the optimal treatment for each patient is complicated by heterogeneous responses driven by diverse clinical and multi-omics characteristics.
Survival prediction with treatment-specific counterfactual reasoning can help clinicians anticipate potential outcomes under different therapies and make more informed, personalized decisions. \textit{Our proposed method does not replace medical judgment but complements it, contributing to more consistent and patient-centered care in real-world healthcare.}

Patients with complex or rapidly progressing diseases, such as cancer, often face severe constraints in time, health, and financial resources, leaving little margin for trial-and-error in treatment selection.
In such contexts, AI systems capable of efficiently retrieving and reasoning over multimodal patient data can assist clinicians by saving diagnostic time, uncovering subtle cross-patient patterns, and supporting more timely evidence-based care.
However, directly retrieving similar patients based on observed features is often unreliable due to limited data, high-dimensional noise, and cohort imbalance.
We instead perform latent subgroup retrieval, implicitly aligning patients through learned treatment–response patterns rather than explicit nearest-neighbor comparisons, thereby enhancing both robustness and interpretability in clinical decision support.

\textbf{Necessary Preconditions.}
%For CURE to achieve its intended impact, several conditions must hold.
Although CURE can mitigate challenges related to censored or incomplete data, hospitals must still provide access to diverse, high-quality multimodal data, including clinical, demographic, and multi-omics information. 
On the other hand, rigorous clinical validation and ethical oversight are essential prior to deployment. The model’s recommendations should be verified through retrospective or prospective studies to ensure consistency with medical guidelines and patient outcomes, under appropriate data governance and patient-consent frameworks. 
Finally, clinicians must remain actively involved in interpretability and decision-review processes. CURE is designed as a decision-support tool, not a replacement for clinical judgment; continuous expert feedback is crucial to maintain trust and ensure responsible use across diverse patient populations.

\textbf{Potential Negative Externalities.}
Potential risks include bias amplification if multimodal data contain systematic disparities across demographic groups. Although the risk of overreliance on model outputs is relatively low, continuous clinical oversight is still necessary.% In addition, retrieval-aware reasoning may inadvertently reveal latent patient subgroups based on proxy variables correlated with sensitive attributes, potentially leading to inequitable treatment outcomes.

\section*{Disclosure of Interest}
The authors have no competing interests.
%In summary, by explicitly analyzing both the necessary preconditions and potential harms, this work aims to promote the responsible development and deployment of retrieval-aware survival models in healthcare.

% \section{Acknowledgments}
\bibliographystyle{splncs04}
\bibliography{main}
%

%\bibliographystyle{bst/sn-mathphys-num}





% LNCS Homepage, \url{http://www.springer.com/lncs}, last accessed 2023/10/25
% \end{thebibliography}

\end{document}